\documentclass{article}

\usepackage[utf8]{inputenc} 
\usepackage[T1]{fontenc}    
\usepackage{hyperref}       
\usepackage{url}            
\usepackage{booktabs}       
\usepackage{amsfonts}       
\usepackage{nicefrac}       
\usepackage{microtype}      

\usepackage{caption}
\usepackage{subcaption}

\usepackage{amsmath}
\usepackage{amssymb}
\usepackage{graphicx}

\title{Training capsules as a routing-weighted \\ product of expert neurons}

\author{%
  Michael Hauser \thanks{Michael Hauser has been supported by a postdoctoral fellowship at the Center for Autism Research in the Children's Hospital of Philadelphia. Any opinions, findings and conclusions or recommendations expressed in this publication are those of the author and do not necessarily reflect the views of the sponsoring agencies.} \\
 Children's Hospital of Philadelphia\\
  \texttt{mikebenh@gmail.com}
}

\date{}

\begin{document}

\maketitle

\begin{abstract}
    Capsules are the multidimensional analogue to scalar neurons in neural networks, and because they are multidimensional, much more complex routing schemes can be used to pass information forward through the network than what can be used in traditional neural networks. This work treats capsules as collections of neurons in a fully connected neural network, where sub-networks connecting capsules are weighted according to the routing coefficients determined by routing by agreement. An energy function is designed to reflect this model, and it follows that capsule networks with dynamic routing can be formulated as a product of expert neurons. By alternating between dynamic routing, which acts to both find subnetworks within the overall network as well as to mix the model distribution, and updating the parameters by the gradient of the contrastive divergence, a bottom-up, unsupervised learning algorithm is constructed for capsule networks with dynamic routing. The model and its training algorithm are qualitatively tested in the generative sense, and is able to produce realistic looking images from standard vision datasets.
\end{abstract}


\section{Introduction}

Capsule networks~\cite{hinton2011transforming} with routing by agreement~\cite{sabour2017dynamic} are a substantial departure from standard feedforward neural networks in the sense of the layer wise process by which they transform the data representation. The McCulloch-Pitts model~\cite{mcculloch1943logical} takes the forward activation of a neuron to be a weighted sum of the scalar-valued activations of the previous layer neurons, where the weighting is determined by the parameter weights. Some deep learning functions break from this model, such as the softmax and max pooling functions, since to which node the softmax and max pooling functions route data depend not only on the previous layer activations, but on the activations of neurons in the current layer as well. Most feedforward deep learning systems, such as the original convolutional LeNet~\cite{lecun1998gradient}, or its larger counterparts such as AlexNet~\cite{krizhevsky2012imagenet}, are such that (i) they transform scalar values from neurons and (ii) the routing functions are either non existent or extremely simple.

Comparatively, capsule networks take the forward activation of a capsule to be vector valued~\cite{hinton2011transforming}. One advantage of a vector valued capsule over a scalar valued neuron is that with a vector one can separate the activation of the capsule from what the capsule is representing. This is done by encoding the activation of the capsule into the magnitude of the vector, and the instantiation parameters for representing the object in the orientation of the vector.

A second advantage of vector valued capsules is that one can develop much more complex routing schemes to pass information forward through the network. With routing by agreement~\cite{sabour2017dynamic}, each vector is linearly mapped to another vector by a learned prediction matrix, and a weighted sum of these predicted vectors becomes the vector to be passed through a nonlinear activation, where the weights of the weighted sum are determined online by routing by agreement. This type of forward pass of information, or routing, is substantially different than the parameter-weighted sum of scalar activations of the McCulloch-Pitts model. It is different because the scalar routing weights determined by the routing by agreement is a complex nonlinear weighting of all of the next layer activations, whereas the McCulloch-Pitts model has no dependence on the next layer activations, the softmax function is a linear weighting of the next layer activations and the max pooling function is a simple nonlinear function of the next layer activations and discards a large amount of information.

Because capsule network models with dynamic routing are substantially different from neural network models in the sense described above, it isn't necessarily straight-forward to extend the learning algorithms that were designed for neural networks to capsule networks. Nevertheless, many of the neural network learning algorithms are quite general and can be extended to capsule networks without much modification, which has led to using backpropagation~\cite{rumelhart1985learning,hinton2011transforming,sabour2017dynamic} and expectation-maximization~\cite{hinton2018matrix} in capsule networks. Similarly, the unsupervised learning method of generative adversarial networks~\cite{goodfellow2014generative} has been extended to capsule networks~\cite{jaiswal2018capsulegan}, but this is built off of backpropagation~\cite{rumelhart1985learning}. 

We build on these efforts by developing an unsupervised learning algorithm for capsule networks with routing by agreement. We do this by considering the capsule network with routing by agreement as a routing-weighted product of expert neurons~\cite{hinton2002training,welling2005exponential}. In this way collections of scalar valued neurons are grouped together to form vector valued capsules, and as capsules we can leverage the routing by agreement algorithm to dynamically route the information.

\section{Algorithm review}

Because the techniques developed in this paper are based substantially off of the mathematics and intuitions from both product of experts learning as well as routing by agreement, we will briefly review these techniques.

\subsection{Product of experts learning}

Product of experts models~\cite{hinton2002training,welling2005exponential,fischer2012introduction} are in a way similar to mixture models, except one replaces the sum of the densities with a product of densities. The intuition is that mixture models act as logical \emph{or} units since each component of the mixture model can alone say if the data belongs to the distribution or not, whereas a product of experts is analogous to logical \emph{and} units, since any of the experts composing the product can exclude the data belonging to the distribution with a low-enough probability.

For models that distinguish between a visible layer and a hidden layer, where connections do not exist within each layer, only across layers, the energy of a certain configuration of visible and hidden binary neurons is defined as follows:

\begin{equation}
E\left(v,h\right) =  -
\sum_{i,j} h_j w_{ij} v_i -
\sum_{i} b_{i} v_i - 
\sum_{j} c_{j} h_j
\end{equation}

With this, the probability for a configuration of visible and hidden units is
$p\left(v,h\right) = \frac{1}{Z} e^{-E\left(v,h\right)}$ where $Z$ is the normalizing partition function. In the product of experts formulation, the density over the visible states, after marginalizing out the binary hidden states $h_j$, is given by

\begin{equation}
    p\left(v\right) =
    \sum_h
    p\left(v,h\right) =
    \frac{1}{Z}
    \sum_h
    e^{-E\left(v,h\right)}
\end{equation}

It follows that the gradient of the log likelihood is given by

\begin{equation}
\frac{\partial \log p}{\partial \Theta} \left(v|\Theta\right) =
- \sum_h p\left(h|v\right) \frac{\partial E}{\partial \Theta} \left(v,h\right) +
\sum_v p\left(v\right) \sum_h p\left(h|v\right) \frac{\partial E}{\partial \Theta} \left(v,h\right) 
\end{equation}

On the right-hand side of this equation, the first term is the expectation of the gradient of the energy, with the expectation being over the true data density. The second term is the expectation of the gradient of the energy, with the expectation being over the data generated by the product of experts model itself. This then leads to the well known learning rule:

\begin{equation}
    \frac{\partial \log p}{\partial w_{ij}} \left(v|\Theta\right) =
    \left\langle v_i h_j \right\rangle_{\textnormal{data}} -
    \left\langle v_i h_j \right\rangle_{\textnormal{model}}
\end{equation}

Finding the model distribution requires mixing with the Markov chain generated by reconstructing the visible and hidden units to infinity. Instead, contrastive divergence~\cite{hinton2002training} minimizes the difference of (i) the KL divergence between the data and the model at infinity with (ii) the one-step Markov mixing and the model at infinity, which means the Markov chain only needs to go through one iteration for each gradient update.

\subsection{Routing by agreement in capsule networks}

As mentioned earlier, capsule networks~\cite{hinton2011transforming} with routing by agreement~\cite{sabour2017dynamic} are a substantial rethinking in the forward flow of information through the network, and this will be briefly reviewed.

At layer $l$, given a collection of vector-valued capsules $ x^{(l)}_{i} $ for $i=1,2,\dots,I$ and matrix-valued prediction maps $W^{(l)}_{ij}$, the pre-activation, vector-valued predicted capsule $j$ from capsule $i$ is $z^{(l+1)}_{j|i}=W^{(l)}_{ij}\cdot x^{(l)}_{i}$. We then take a weighted average of all of the predictions made at layer $l$ to yield the final, pre-activation, vector-valued capsule $j$ at layer $l+1$, i.e.
\begin{equation}
z^{(l+1)}_{j}=\sum_{i}c^{(l)}_{ij} W^{(l)}_{ij}\cdot x^{(l)}_{i}
\end{equation}
where the scalar-valued $c^{(l)}_{ij}$'s are determined by routing by agreement and $\sum_{i}c^{(l)}_{ij}=1$. This is done for all capsules $j=1,2,\dots,J$ at layer $l+1$. 

Routing by agreement is an iterative procedure to re-weigh each of the individual predictions to yield a final prediction, of a similar flavor to boosting weak learners to yield a strong learner. If an individual prediction $z^{(l+1)}_{j|i}$ agrees well with the collective prediction $z^{(l+1)}_{j}$, then we would like to increase the $i$ to $j$ routing weight $c^{(l)}_{ij}$, whereas if they disagree we would like to decrease that same routing weight. Agreement has been measured as both the inner product~\cite{sabour2017dynamic}, as well as the cosine distance~\cite{hinton2018matrix}, between the individual predictions and the squashed collective prediction. In this work we use the cosine distance.

The squashing function we use, from~\cite{sabour2017dynamic}, is defined as follows:

\begin{equation}
    \textnormal{squash}\left(z^{(l+1)}_{j}) \right)=
    \frac{\Vert z^{(l+1)}_{j}\Vert^2}{1+\Vert z^{(l+1)}_{j}\Vert^2} 
    \frac{z^{(l+1)}_{j}}{\Vert z^{(l+1)}_{j}\Vert} 
\end{equation}

The intention of the squashing function is to scale the magnitude of $z^{(l+1)}_{j}$ between $0$ and $1$ while keeping the orientation the same. We can then interpret capsule $j$ is on if the magnitude of the squashing function is close to $1$, and $j$ is off if the magnitude is close to $0$:
\begin{equation}
    P\left(j=\textnormal{on}|x^{(l)}_1,\dots,x^{(l)}_{I}\right) = 
    \Vert \textnormal{squash}\left(z^{(l+1)}_{j} \right) \Vert
\end{equation}

Additionally, since the orientation is decoupled from the magnitude, a single capsule can fire over the entire range of input orientations, as compared to a neuron which can only fire under its one specific oriented input. These design feature allows the orientation to encode the instantiation parameters of the object that the capsule is representing, while the magnitude encodes whether the capsule is on or off. With a scalar valued neuron it is only possible to encode whether the neuron is on or off.

\begin{figure}[t]
     \centering
         \centering
        \includegraphics[width=\textwidth]{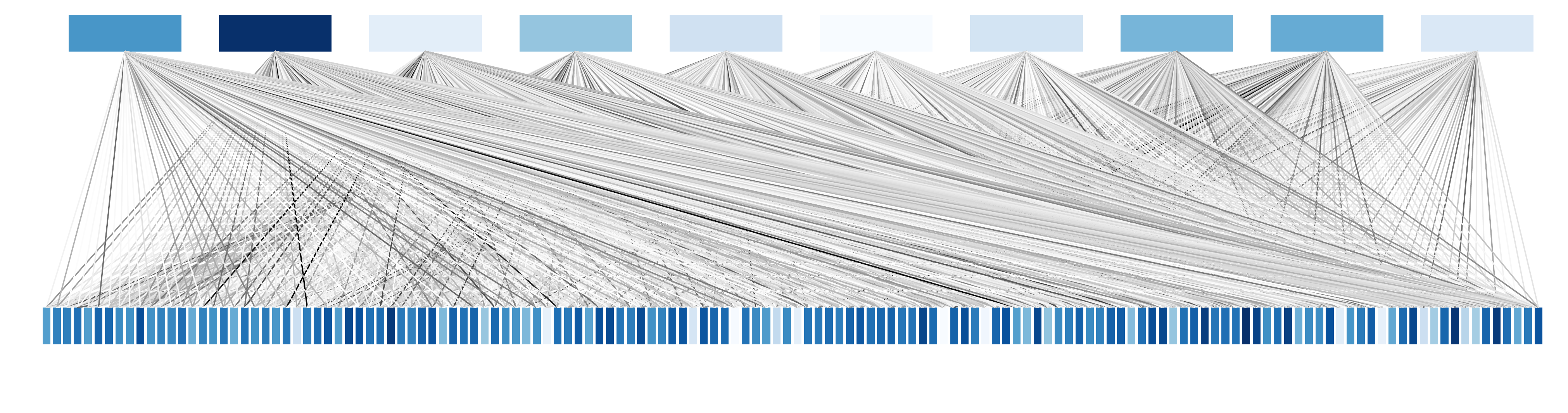}
         \caption{Routing diagram between layers with $144$ capsules at layer $l$ and $10$ capsules at layer $l+1$. The rectangles represent capsules, and are color coded so that lighter colors represent low-probability activations while darker colors represent high-probability activations. Similarly, the edges connecting the rectangles are the routing coefficients $c^{(l)}_{ij}$'s, and for lighter edges there is a lower routing weight, while for darker edges there is a higher routing weight. This capsule network was trained on MNIST in the unsupervised way described in this paper.}
    \label{fig:routing-diagram}
\end{figure}

To aid in visualizing the routing process, Figure~\ref{fig:routing-diagram} was made using the unsupervised learning algorithm developed in this paper. The $144$ capsules on the first layer are routed to $10$ capsules on the second layer. Each of the first layer capsules are composed of $8$ neurons while each of the second layer capsules are composed of $16$ neurons. If the capsule has a magnitude close to $0$ it is a lighter shade, while if it its magnitude is close to $1$ it is a darker shade. Similarly the edge connections are the scalar valued routing weights $c^{(l)}_{ij}$'s, where light edges correspond to a weak routing weight while a dark edge corresponds to a strong routing weight. As a neural network each edge would be replaced by a fully connected $8\times 16$ matrix of edges for each of the matrix valued prediction maps $W^{(l)}_{ij}$.

\section{Capsule networks as a product of expert neurons }

This section develops the product of experts formulation of capsule networks. The model treats capsules as collections of neurons, and weights these collections of neurons across layers according to the dynamic routing between capsules algorithm.

Our intuition is that during optimization, alternating between routing by agreement and updating parameters, the model will be able to better learn to leverage the routing by agreement mechanism. This is how capsule networks with routing by agreement are trained with backpropagation, and so we are extending this intuition to product of experts learning. In both cases, the optimization procedure is a two step process.

\begin{enumerate}
    \item
    With $W^{(l)}_{ij}$'s fixed, perform routing by agreement to determine the routing weights $c^{(l)}_{ij}$'s
    \item
    With $c^{(l)}_{ij}$'s fixed, define the energy and update the parameter weights $W^{(l)}_{ij} \xleftarrow[]{} W^{(l)}_{ij} + \Delta W^{(l)}_{ij} $
\end{enumerate}

For clarity, we sometimes write the individual, vector-valued capsule $x^{(l)}_{i}$ and $x^{(l+1)}_{j}$ in terms of the scalar-valued neuron components $x^{(l)}_{i,m}$ and $x^{(l+1)}_{j,n}$, as $x^{(l)}_{i}=[x^{(l)}_{i,1};x^{(l)}_{i,2};\dots;x^{(l)}_{i,M}]$ for an $M$-dimensional capsule at layer $l$ and $x^{(l+1)}_{j}=[x^{(l+1)}_{j,1};x^{(l+1)}_{j,2};\dots;x^{(l+1)}_{j,N}]$ for an $N$-dimensional capsule at layer $l+1$. Similarly, we write the entire collection of capsules in layer $l$ as $x^{(l)}=[x^{(l)}_{1},x^{(l)}_{2},\dots,x^{(l)}_{I}]$ for $I$-many capsules in layer $l$, and $x^{(l+1)}=[x^{(l+1)}_{1},x^{(l+1)}_{2},\dots,x^{(l+1)}_{J}]$ for $J$-many capsules in layer $l+1$. We use $i$ and $m$ to index over layer $l$ and $j$ and $n$ to index over layer $l+1$, and we reserve the dot symbol $\cdot$ for matrix-vector multiplication, as in $W \cdot x$.

\subsection{Routing-Weighted Product of Expert Neurons}
\label{sec:routing-weighted-poen}

Define the energy between vector-valued capsules $x^{(l)}_i$ and $x^{(l+1)}_j$

\begin{equation}
E_{ij}\left(x^{(l)}_{i},x^{(l+1)}_{j}\right)=
-
x^{(l+1)T}_{j} \cdot W^{(l)}_{ij} \cdot x^{(l)}_{i}
\end{equation}

With this, the total energy between layer $l$ and $l+1$ is the sum over all possible combinations of these energies between capsules, weighted by the scalar routing weights $c^{(l)}_{ij}$'s that are determined by routing by agreement.

\begin{equation}
    E\left(x^{(l)},x^{(l+1)}\right) = 
    \sum_{i,j} 
    c^{(l)}_{ij}
    E_{ij}\left(x^{(l)}_{i},x^{(l+1)}_{j}\right)
\end{equation}

Defined in this way, the significance of the routing coefficients $c^{(l)}_{ij}$'s act to weigh the energy between capsules $i$ and $j$. If a capsule is understood as being a collection of neurons, then this is a routing-weighted energy between the $i^{th}$ collection of neurons at layer $l$ and the $j^{th}$ collection of neurons at layer $l+1$, given by $ E_{ij}\left(x^{(l)}_{i},x^{(l+1)}_{j}\right) =  \sum_{m,n} 
 x^{(l+1)}_{j,n} W^{(l)}_{ij,mn} x^{(l)}_{i,m} $.

The intuition is that, between capsules $i$ and $j$, the routing coefficients will increase the energy of the sub-networks that route to each other, and decrease the energy of the sub-networks that do not route to each other.

Following the scalar product of experts formulation, we have the total probability is

\begin{equation}
    P\left(x^{(l)},x^{(l+1)}\right)=
    \frac{1}{Z}
    e^{-E\left(x^{(l)},x^{(l+1)}\right)}
\end{equation}
where $Z$ is the normalizing partition function. This can be written as a routing-weighted product of expert neurons:

\begin{equation}
\label{eqn:routing-weighted-product-of-expert-neurons}
P\left(x^{(l)},x^{(l+1)}\right)=
\frac{1}{Z}
\Pi_j \Pi_n
\exp\left(
    \sum_{i} 
c^{(l)}_{ij}
\sum_{m} 
x^{(l+1)}_{j,n} W^{(l)}_{ij,mn} x^{(l)}_{i,m}
\right)    
\end{equation}

In vector notation, Equation~\ref{eqn:routing-weighted-product-of-expert-neurons} is equivalent to

\begin{equation}
    P\left(x^{(l)},x^{(l+1)}\right)=
    \frac{1}{Z}
    \Pi_j
    \exp{\left(
        \sum_{i} 
    c^{(l)}_{ij}
    x^{(l+1)T}_{j} \cdot W^{(l)}_{ij} \cdot x^{(l)}_{i}
    \right)}
\end{equation}

Vector notation has the advantage that it makes clear the capsule organization of the collections of neurons, and how it relates to routing by agreement. We marginalize over the hidden states with $P\left(x^{(l)}\right)=\sum_{x^{(l+1)}} P\left(x^{(l)},x^{(l+1)}\right) $ to yield the visible state distribution:

\begin{equation}
    P\left(x^{(l)}\right)=
    \frac{1}{Z}
    \Pi_j
    \left[
    1+
    \exp{\left(
        \sum_{i} 
    c^{(l)}_{ij}
    W^{(l)}_{ij} \cdot x^{(l)}_{i}
    \right)}
    \right]
\end{equation}

We can now perform inference for Gibb's sampling as $ P\left(x^{(l+1)}_{j,n}=1 | x^{(l)} \right)=
    \frac{1}
    {1+
    \exp\left(-
        \sum_{i} 
    c^{(l)}_{ij}
    \sum_{m} 
    W^{(l)}_{ij,mn} x^{(l)}_{i,m}
    \right)} $, or in vector notation:

\begin{equation}
\label{eqn:inference_forward}
    P\left(x^{(l+1)}_{j}=1 | x^{(l)} \right)=
    \frac{1}
    {1+
    \exp\left(-
        \sum_{i} 
    c^{(l)}_{ij}
    W^{(l)}_{ij} 
    \cdot
    x^{(l)}_{i}
    \right)}
\end{equation}

In the reverse direction the analogous equation is:

\begin{equation}
\label{eqn:inference_reverse}
    P\left(x^{(l)}_{i}=1 | x^{(l+1)} \right)=
    \frac{1}
    {1+
    \exp\left(-
        \sum_{j} 
    c^{(l)}_{ij}
    W^{(l)T}_{ij}
    \cdot
    x^{(l+1)}_{j}
    \right)}    
\end{equation}
where $W^{(l)T}_{ij}$ is the matrix-transpose of $W^{(l)}_{ij}$, and is not the same thing as $W^{(l)}_{ji}$.

Subtleties regarding Equations~\ref{eqn:inference_forward} and ~\ref{eqn:inference_reverse} deserve discussion. In routing by agreement, the individual capsule $i$ makes its prediction about capsule $j$ as $z^{(l+1)}_{j|i}=W^{(l)}_{ij} \cdot x^{(l)}_{i}$, and then the routing-weighted average with the other predictions about $j$ from each of the $i$'s is taken, yielding $z^{(l+1)}_{j}=\sum_i c^{(l)}_{ij} z^{(l+1)}_{j|i}$, where the weighted average is over the routing weights $c^{(l)}_{ij}$'s such that $\sum_i c^{(l)}_{ij}=1 $.

In this way, Equation~\ref{eqn:inference_forward} is the sigmoid activation of this routing-weighted capsule prediction, where the sigmoid nonlinearity acts element-wise, and the value of each dimension is independent of the values of the other dimensions. This is in comparison to the squashing nonlinearity used during the routing by agreement, where the values of each dimension are dependent on the values of the other dimensions, to ensure that the vector is only scaled without changing its orientation.

Interpretting Equation~\ref{eqn:inference_reverse} in a similar way suggests that capsule $x^{(l+1)}_{j}$ is making its prediction for the layer below as $z^{(l)}_{i|j}=W^{(l)T}_{ij} \cdot x^{(l+1)}_{j}$, and then keeping the same routing weights $c^{(l)}_{ij}$'s to give $z^{(l)}_{i}=\sum_j c^{(l)}_{ij} z^{(l)}_{i|j}$. This is not a proper weighted average since $\sum_i c^{(l)}_{ij}=1 $, whereas the weighted sum is over the $j$'s, as opposed to the $i$'s. It is not straightforward to use a different set of routing weights $e^{(l+1)}_{ji}$ such that $\sum_j e^{(l+1)}_{ji}=1 $, because this would mean having a different energy function for the forwards and backwards passes. 

Since we divide the optimization procedure into two steps these are not issues, just as is the case when using backpropagation with routing by agreement. The first step uses the capsule structure of the network to perform routing by agreement in order to decide how to weigh pairs of capsules $i$ and $j$. The second step ignores the capsule structure of the network, except for the fact that it weighs the $i^{th}$ group of neurons in layer $l$ with the $j^{th}$ group of neurons in layer $l+1$ by $c^{(l)}_{ij}$. In this way Equations~\ref{eqn:inference_forward} and~\ref{eqn:inference_reverse} can be viewed as groups of neurons in one layer inferring activations about groups of neurons in another layer.

\subsection{Gradient of the log-likelihood}

To optimize the log-likelihood, we take the following gradient:

\begin{multline}
    \frac{\partial}{\partial W^{(l)}_{ij}}
    \log P\left(x^{(l)}\right) = 
    \sum_{x^{(l+1)}_j }
    P\left( x^{(l+1)}_j | x^{(l)}_i \right) 
    \frac{\partial E_{ij}}{\partial W^{(l)}_{ij}}\left(x^{(l)}_i,x^{(l+1)}_j\right) 
    \\
    -
    \sum_{x^{(l)}_i }
    P\left( x^{(l)}_i \right)
    \sum_{x^{(l+1)}_j }
    P\left( x^{(l+1)}_j | x^{(l)}_i \right) 
    \frac{\partial E_{ij}}{\partial W^{(l)}_{ij}}\left(x^{(l)}_i,x^{(l+1)}_j\right) 
\end{multline}
where we used the fact that $ \frac{\partial E}{\partial W^{(l)}_{ij}} = \sum_{i'j'} \frac{\partial E_{i'j'}}{\partial W^{(l)}_{ij}} = \frac{\partial E_{ij}}{\partial W^{(l)}_{ij}} $. The gradient of the energy has a simple form:

\begin{equation}
\frac{\partial E_{ij}}{\partial W^{(l)}_{ij} }\left(x^{(l)},x^{(l+1)}\right) = 
    c^{(l)}_{ij}
    x^{(l+1) T}_{j} x^{(l)}_{i}
\end{equation}

We now have the update rule for the parameter weights:
\begin{equation}
    \frac{\partial}{\partial W^{(l)}_{ij}}
    \log P\left(x^{(l)}\right) = 
    c^{(l)}_{ij} \left(
    \langle x^{(l)}_{i} x^{(l+1)}_{j} \rangle_{\textnormal{data}}
    -
    \langle x^{(l)}_{i} x^{(l+1)}_{j} \rangle_{\textnormal{model}}
    \right)
\end{equation}

We see that without routing by agreement, this update rule would be identical to the RBM update rule. With routing by agreement, this tells us that the parameters are updated proportionally to the routing weights $c^{(l)}_{ij}$'s. This means that the routing weights appear in two ways, one in determining the forward/backward sampling of $x^{(l+1)}_j = \sigma \left( \sum_i c^{(l)}_{ij}W^{(l)}_{ij}\cdot x^{(l)}_i \right)$, and second in proportionally weighing the parameter updates.

As is well known, estimating the model distribution requires sampling the Markov chain to infinity. Instead we minimize the contrastive divergence~\cite{hinton2002training}, and so we only run the Markov chain for a single iteration.

\section{Experiments}

\begin{figure}[t]
    \centering
    \includegraphics[width=\textwidth]{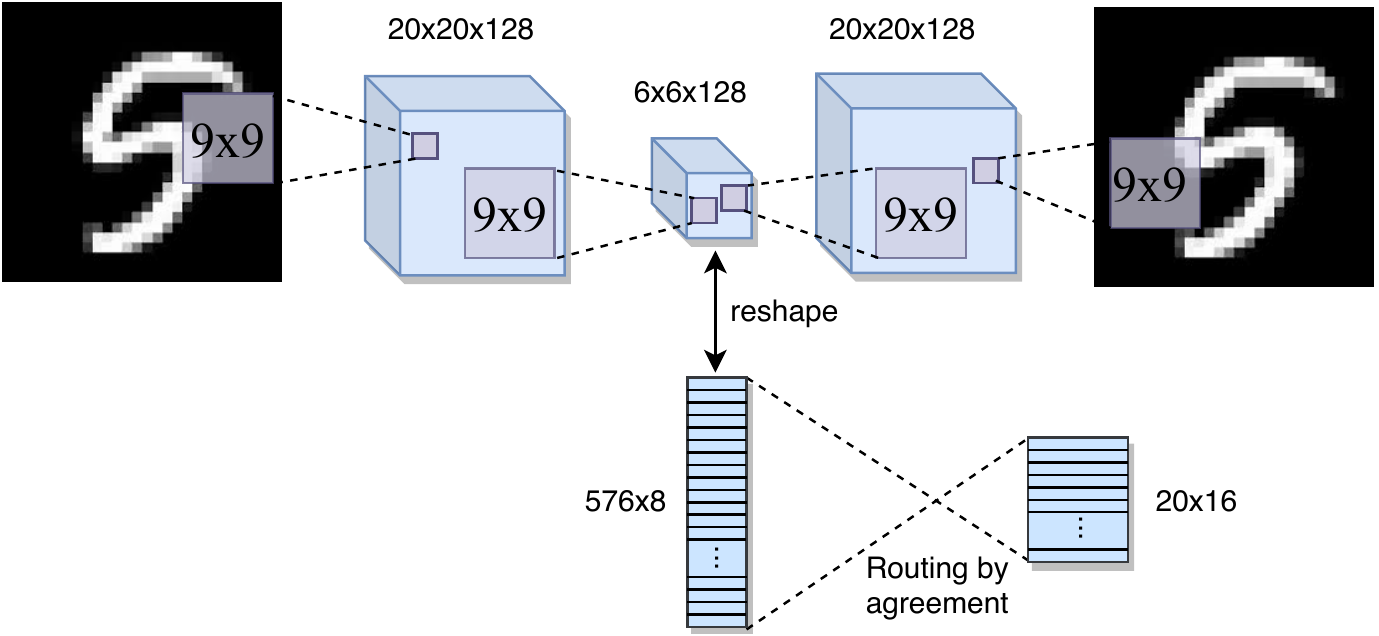}
    \caption{The network architecture used for the experiments. First the autoencoding convolutional network is trained to learn the filter weights. Once this is complete, those weights are held fixed and we train the capsule network as a product of expert neurons by feeding the data through the convolutional encoder. Finally for testing, we randomly sample points from a $16$-dimensional, zero mean, unit variance Gaussian, run these points through routing by agreement in reverse, and then finally through the convolutional decoder to produce the image.}
    \label{fig:network-diagram}
\end{figure}

This section details the experiments conducted, training the capsule networks in the unsupervised ways described in the preceding section. The network diagram can be seen in Figure~\ref{fig:network-diagram}. In all experiments, we use an autoencoding convolutional neural network~\cite{zeiler2010deconvolutional}, with dropout~\cite{srivastava2014dropout}, to learn the first two layers of filters in an unsupervised way. We denote filter bank sizes by $ \left[ \textnormal{height}, \textnormal{width}, \textnormal{channels-in}, \textnormal{channels-out} \right] $. The first filter bank is of size $\left[9,9,1,128\right]$ for grey-scale or $\left[9,9,3,128\right]$ for rgb, followed by a Leaky ReLU activation, while the second filter bank is $\left[9,9,128,128\right]$ followed by a sigmoid activation. We use a sigmoid for the second hidden layer activation because we want to map the data between $0$ and $1$ to match the energy model's sigmoids. For the decoder, first we use a Leaky Relu, and then a sigmoid; this sigmoid is to map the pixel intensities between $0$ and $1$.

Once the autoencoder has been sufficiently trained, we fix these filter weights and next train the capsule network as a product of expert neurons. The $128$-channel hidden layer is reshaped to $6\times 6\times 128/8=576$ capsules, each having $8$ dimensions. These are then mapped to $20$ capsules, each having $16$ dimensions, by $x^{(l+1)}_j = \sigma \left( \sum_i c^{(l)}_{ij}W^{(l)}_{ij}\cdot x^{(l)}_i \right)$, and in reverse $ x^{(l)}_i = \sigma \left( \sum_j c^{(l)}_{ij}W^{(l)T}_{ij}\cdot x^{(l+1)}_j \right)$ with the $c^{(l)}_{ij}$'s fixed from the forward pass of the routing by agreement. All updates use sgd with momentum, learning rate decay, an $\ell_2$ regularization on the weights and is implemented in TensorFlow~\cite{abadi2016tensorflow}. On a single gpu it takes about 30-60 minutes to train, depending on the dataset.

\subsection{Routing-Weighted Product of Expert Neurons}

\begin{figure}[]
     \centering
     \begin{subfigure}[b]{1.\textwidth}
         \centering
        \includegraphics[width=\textwidth]{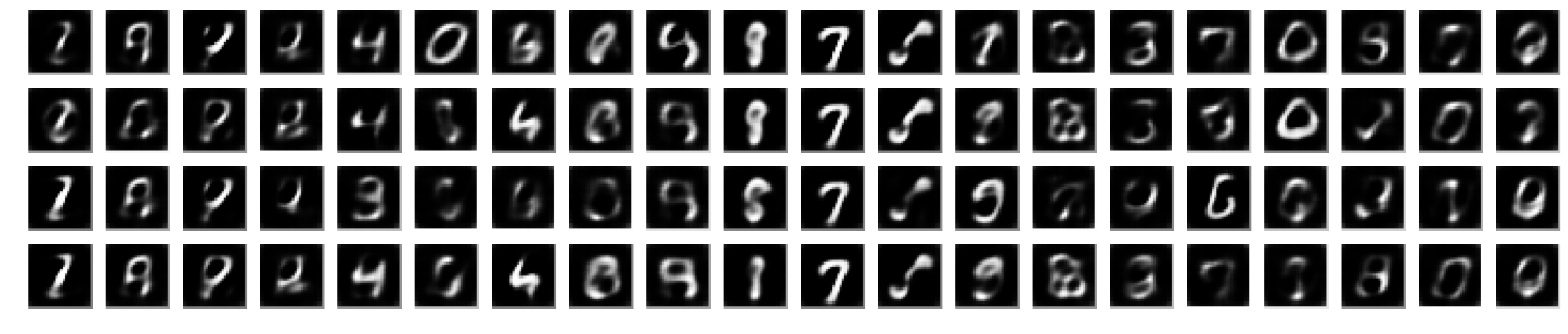}
         \caption{MNIST}
         \label{fig:y equals x}
     \end{subfigure}
     \hfill
     \begin{subfigure}[]{1.\textwidth}
         \centering
        \includegraphics[width=\textwidth]{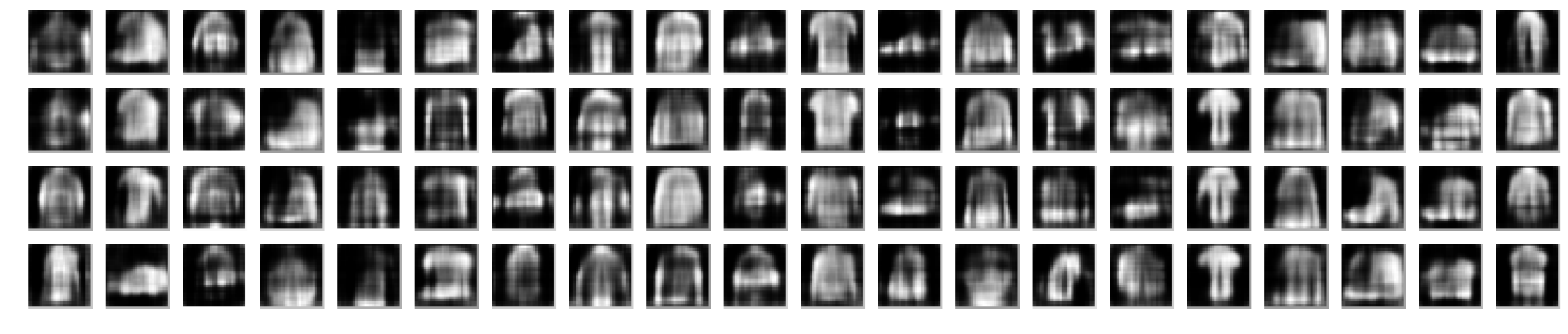}
         \caption{Fashion-MNIST}
         \label{fig:three sin x}
     \end{subfigure}
     \hfill
     \begin{subfigure}[b]{1.\textwidth}
         \centering
         \includegraphics[width=\textwidth]{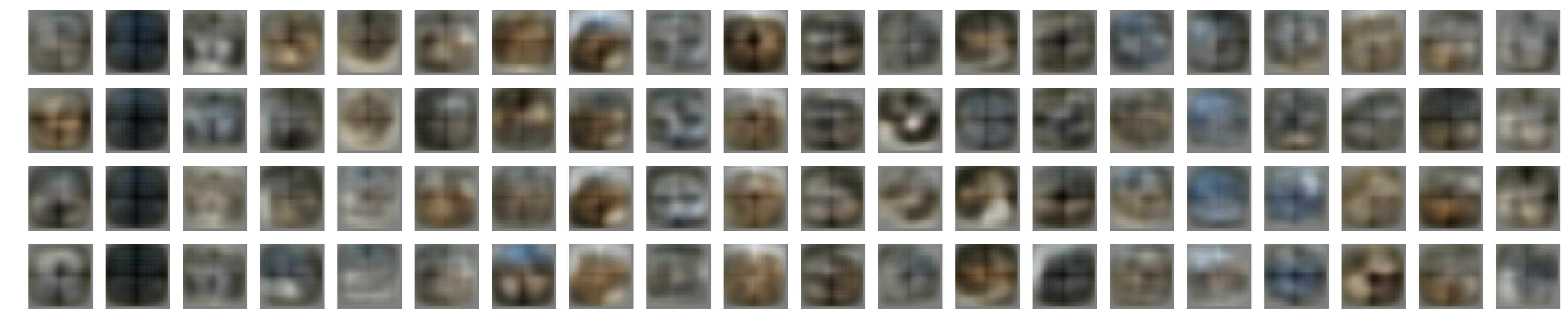}
         \caption{CIFAR10}
         \label{fig:five over x}
     \end{subfigure}
    \caption{Images created by the unsupervised, routing-weighted product of expert neurons model. For each model, all of the 80 images were sampled together so these are not cherry picked examples. Each column is one of the twenty hidden layer capsules, while the four rows are four random samplings for that $16$-dimensional capsule. It is seen that individual capsules learn specific objects in this unsupervised setting, and different samples drawn from these capsules, i.e. different instatiation parameters, yield changes in the object. For example in MNIST these instantiation dimensions yield changes in stroke thickness, angles and lengths, while in Fashion-MNIST they can transform dimensions such as sleeve length.}
    \label{fig:routing-weighted-poen}
\end{figure}

Results generated from the routing-weighted product of expert neurons model of Section~\ref{sec:routing-weighted-poen} can be seen in Figure~\ref{fig:routing-weighted-poen}. For each of the $20$ hidden layer capsules, four points are randomly sampled from a $16$-dimensional, zero mean, unit variance Gaussian, and then fed into a sigmoid function. All hidden layer capsules are set to $0$ except for one, which is assigned this random sigmoided $16$-dimensional point. The network is then run in reverse, from the hidden layer to the input, where capsules are routed according to $z^{(l)}_i = \sum_j c^{(l)}_{ij} W^{(l) T}_{ij} \cdot x^{(l+1)}_j $, where the $c^{(l)}_{ij}$'s are determined by routing by agreement, but are normalized along the $l^{th}$-layer, i.e. $\sum_i c^{(l)}_{ij}=1$. In this manner, this then acts as an unsupervised, generative model, creating the $4 \times 20$ images of Figure~\ref{fig:routing-weighted-poen}. It is noted that these $4 \times 20$ random points were sampled together from the Gaussian, and so these are not cherry picked examples.

Each capsule seems to learn specific objects in this unsupervised setting. For example on MNIST, capsules seems to learn digits of different widths, angles and thicknesses. Nearly all of the capsules seem to have dimensions that correspond to the thickness of the digit. Interestingly, other dimensions are sometimes seen to extend strokes in different ways. For example, one of the capsules extends $9$'s to $8$'s by adding a stroke in the bottom left, as well as extends $9$'s to $4$'s by eliminating the stroke on the top.

\section{Conclusions}

This work develops a formulation of capsule networks as a weighted product of expert neurons, where the energy connecting two capsules is weighted proportionally by the routing coefficients from the routing by agreement mechanism. This contributes to the effort of further developing capsule networks by developing a new unsupervised training algorithm specifically designed for capsule networks that can build capsule networks from the ground up in an unsupervised way. Initial results on MNIST, Fashion-MNIST and CIFAR suggest that this learning procedure retains the attractive attributes of capsule networks, namely their ability to represent an object's instantiation parameters in the orientation of the vector, in the developed unsupervised setting.

\newpage


\begin{thebibliography}{10}

\bibitem{hinton2011transforming}
Geoffrey~E Hinton, Alex Krizhevsky, and Sida~D Wang.
\newblock Transforming auto-encoders.
\newblock In {\em International Conference on Artificial Neural Networks},
  pages 44--51. Springer, 2011.

\bibitem{sabour2017dynamic}
Sara Sabour, Nicholas Frosst, and Geoffrey~E Hinton.
\newblock Dynamic routing between capsules.
\newblock In {\em Advances in neural information processing systems}, pages
  3856--3866, 2017.

\bibitem{mcculloch1943logical}
Warren~S McCulloch and Walter Pitts.
\newblock A logical calculus of the ideas immanent in nervous activity.
\newblock {\em The bulletin of mathematical biophysics}, 5(4):115--133, 1943.

\bibitem{lecun1998gradient}
Yann LeCun, L{\'e}on Bottou, Yoshua Bengio, Patrick Haffner, et~al.
\newblock Gradient-based learning applied to document recognition.
\newblock {\em Proceedings of the IEEE}, 86(11):2278--2324, 1998.

\bibitem{krizhevsky2012imagenet}
Alex Krizhevsky, Ilya Sutskever, and Geoffrey~E Hinton.
\newblock Imagenet classification with deep convolutional neural networks.
\newblock In {\em Advances in neural information processing systems}, pages
  1097--1105, 2012.

\bibitem{rumelhart1985learning}
David~E Rumelhart, Geoffrey~E Hinton, and Ronald~J Williams.
\newblock Learning internal representations by error propagation.
\newblock Technical report, California Univ San Diego La Jolla Inst for
  Cognitive Science, 1985.

\bibitem{hinton2018matrix}
Geoffrey~E Hinton, Sara Sabour, and Nicholas Frosst.
\newblock Matrix capsules with em routing.
\newblock 2018.

\bibitem{goodfellow2014generative}
Ian Goodfellow, Jean Pouget-Abadie, Mehdi Mirza, Bing Xu, David Warde-Farley,
  Sherjil Ozair, Aaron Courville, and Yoshua Bengio.
\newblock Generative adversarial nets.
\newblock In {\em Advances in neural information processing systems}, pages
  2672--2680, 2014.

\bibitem{jaiswal2018capsulegan}
Ayush Jaiswal, Wael AbdAlmageed, Yue Wu, and Premkumar Natarajan.
\newblock Capsulegan: Generative adversarial capsule network.
\newblock In {\em Proceedings of the European Conference on Computer Vision
  (ECCV)}, pages 0--0, 2018.

\bibitem{hinton2002training}
Geoffrey~E Hinton.
\newblock Training products of experts by minimizing contrastive divergence.
\newblock {\em Neural computation}, 14(8):1771--1800, 2002.

\bibitem{welling2005exponential}
Max Welling, Michal Rosen-Zvi, and Geoffrey~E Hinton.
\newblock Exponential family harmoniums with an application to information
  retrieval.
\newblock In {\em Advances in neural information processing systems}, pages
  1481--1488, 2005.

\bibitem{fischer2012introduction}
Asja Fischer and Christian Igel.
\newblock An introduction to restricted boltzmann machines.
\newblock In {\em iberoamerican congress on pattern recognition}, pages 14--36.
  Springer, 2012.

\bibitem{zeiler2010deconvolutional}
Matthew~D Zeiler, Dilip Krishnan, Graham~W Taylor, and Robert Fergus.
\newblock Deconvolutional networks.
\newblock In {\em Cvpr}, volume~10, page~7, 2010.

\bibitem{srivastava2014dropout}
Nitish Srivastava, Geoffrey Hinton, Alex Krizhevsky, Ilya Sutskever, and Ruslan
  Salakhutdinov.
\newblock Dropout: a simple way to prevent neural networks from overfitting.
\newblock {\em The Journal of Machine Learning Research}, 15(1):1929--1958,
  2014.

\bibitem{abadi2016tensorflow}
Mart{\'\i}n Abadi, Paul Barham, Jianmin Chen, Zhifeng Chen, Andy Davis, Jeffrey
  Dean, Matthieu Devin, Sanjay Ghemawat, Geoffrey Irving, Michael Isard, et~al.
\newblock Tensorflow: A system for large-scale machine learning.
\newblock In {\em 12th $\{$USENIX$\}$ Symposium on Operating Systems Design and
  Implementation ($\{$OSDI$\}$ 16)}, pages 265--283, 2016.

\end{thebibliography}

\end{document}